\documentclass{article}

\usepackage{microtype}
\usepackage{graphicx}
\usepackage{subcaption}
\usepackage{booktabs} % for professional tables
\usepackage{multirow}
\usepackage{float}
\usepackage{placeins}
\usepackage{enumitem}
\usepackage[svgnames]{xcolor}

\usepackage{hyperref}

\usepackage[preprint]{icml2026}

\usepackage{amsmath}
\usepackage{amssymb}
\usepackage{mathtools}
\usepackage{amsthm}
\usepackage{algorithm}
\usepackage{algorithmic}
\usepackage[capitalize,noabbrev]{cleveref}

\theoremstyle{plain}

\theoremstyle{definition}

\theoremstyle{remark}

\usepackage[textsize=tiny]{todonotes}

\icmltitlerunning{Double-P: Hierarchical Top-P Sparse Attention for Long-Context LLMs}

\begin{document}

\twocolumn[
  \icmltitle{Double-P: Hierarchical Top-P Sparse Attention for Long-Context LLMs}



  \icmlsetsymbol{equal}{*}

  \begin{icmlauthorlist}
    \icmlauthor{Wentao Ni}{ucsd}
    \icmlauthor{Kangqi Zhang}{umich}
    \icmlauthor{Zhongming Yu}{ucsd}
    \icmlauthor{Oren Nelson}{ucsd}
    \icmlauthor{Mingu Lee}{qualcomm}
    \icmlauthor{Hong Cai}{qualcomm}
    \icmlauthor{Fatih Porikli}{qualcomm}
    \icmlauthor{Jongryool Kim}{sk}
    \icmlauthor{Zhijian Liu}{ucsd,nvidia}
    \icmlauthor{Jishen Zhao}{ucsd}
  \end{icmlauthorlist}

  \icmlaffiliation{ucsd}{Department of Computer Science and Engineering, University of California San Diego, La Jolla, United States}
  \icmlaffiliation{umich}{Department of Electrical Engineering and Computer Science, University of Michigan, Ann Arbor, United States}
  \icmlaffiliation{qualcomm}{Qualcomm AI Research, San Diego, United States}
  \icmlaffiliation{sk}{SK hynix America, San Jose, United States}
  \icmlaffiliation{nvidia}{NVIDIA, Santa Clara, United States}
  

  \icmlcorrespondingauthor{Jishen Zhao}{jzhao@ucsd.edu}

  \icmlkeywords{Machine Learning, ICML}

  \vskip 0.3in
]



\printAffiliationsAndNotice{}  % no special notice (required even if empty)

\begin{abstract}

% \begin{itemize}
%     \item Problem: Top-k sparse attention is budget-driven and suboptimal; top-p is accuracy-driven but expensive.
%     \item Key challenge: Accurate attention mass estimation at low cost.
%     \item Insight: Attention mass is hierarchically concentrated (cluster-level → token-level).
%     \item Contribution: First Top-P on cluster centroids to estimate attention mass. Second Top-P to adaptively decide which clusters require exact token attention. Efficient GPU implementation with fused gather + weighted FlashAttention.
%     \item Results: Same accuracy with significantly lower latency vs Twilight; robust across models, tasks, and context lengths.
% \end{itemize}

As long-context inference becomes central to large language models (LLMs), attention over growing key-value caches emerges as a dominant decoding bottleneck, motivating sparse attention for scalable inference. Fixed-budget top-k sparse attention cannot adapt to heterogeneous attention distributions across heads and layers, whereas top-p sparse attention directly preserves attention mass and provides stronger accuracy guarantees. Existing top-p methods, however, fail to jointly optimize top-p accuracy, selection overhead, and sparse attention cost, which limits their overall efficiency.
We present \textbf{Double-P}, a hierarchical sparse attention framework that optimizes all three stages. Double-P first performs coarse-grained top-p estimation at the cluster level using size-weighted centroids, then adaptively refines computation through a second top-p stage that allocates token-level attention only when needed. Across long-context benchmarks, Double-P consistently achieves near-zero accuracy drop, reducing attention computation overhead by up to 1.8$\times$ and delivers up to 1.3$\times$ end-to-end decoding speedup over state-of-the-art fixed-budget sparse attention methods.
% while preserving attention value guarantees. 

\end{abstract}

\section{Introduction}

Recent large language models (LLMs) 
%increasingly support long-context 
support increasingly long context inference, enabling applications such as retrieval-augmented generation \cite{fan2024survey, jin2024long}, document understanding \cite{appalaraju2021docformer, luo2024layoutllm}, and long-horizon reasoning over tens or even hundreds of thousands of tokens \cite{grattafiori2024llama, yang2025qwen2, team2023gemini}. While extended context improves modeling capacity, it also introduces substantial computational and memory overhead during autoregressive decoding due to 
%because of 
attention mechanism. In practice, the attention computation dominates the 
%becomes the dominant 
bottleneck in long-context inference, 
%motivating the need 
calling for more efficient attention mechanisms.

\begin{figure}
    \centering
    \includegraphics[width=0.9\linewidth]{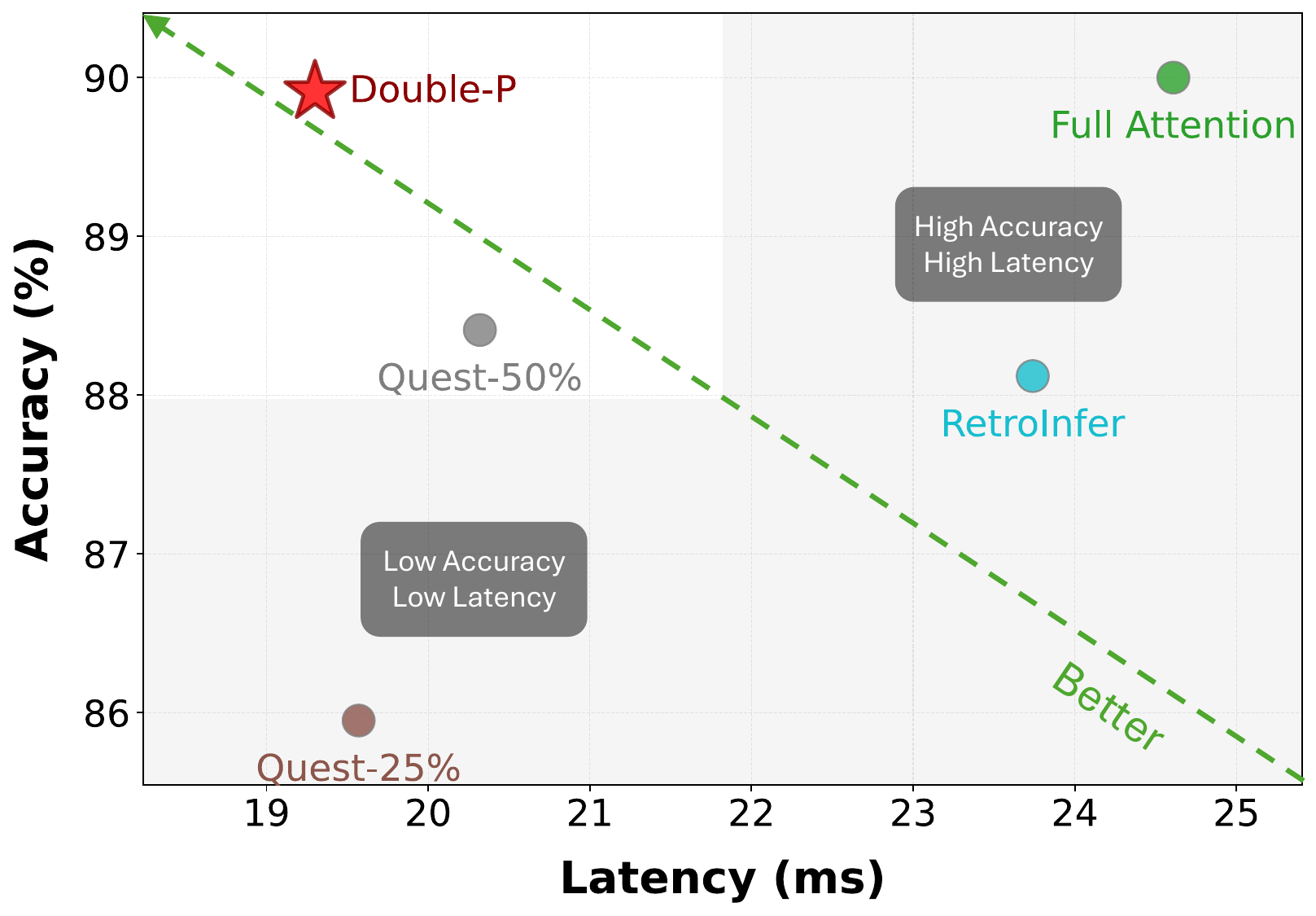}
    \caption{Average accuracy and decode latency of different sparse attention methods tested on Ruler \cite{hsieh2024ruler} 32k context length. Double-P dominates existing approaches, forming a Pareto frontier that demonstrates superior efficiency–accuracy balance.}
    \label{fig:latency_vs_accuracy}
\end{figure}

% \subsection{Sparse Attention}

Sparse attention addresses this bottleneck by restricting attention computation to a subset of tokens that are estimated to be most relevant to the current query. Most existing methods \cite{tang2024quest, liu2024retrievalattention, zhang2025pqcache, liu2025clusterkv, chen2025retroinfer} adopt a \emph{top-k} formulation, where a fixed number of tokens are selected under a predefined budget.

\begin{figure}[t]
    \centering
    \begin{subfigure}{\linewidth}
        \includegraphics[width=\linewidth]{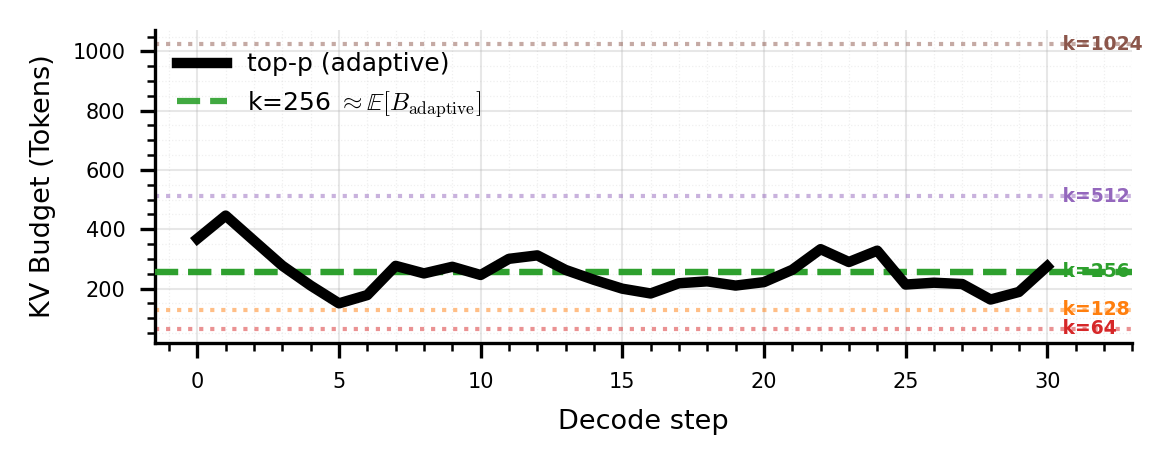}
        % \caption{Adaptive demand vs. static supply (NIAH 8K). The black curve tracks the adaptive token budget required to maintain a captured attention mass of $p=0.95$ at each decode step (averaged across heads and layers). In contrast, the horizontal lines represent static fixed-k budgets that do not accommodate steps where the attention mechanism requires more or less context.} 
        \caption{Adaptive vs. fixed token budgets. The black curve shows the adaptive token budget required to maintain an attention mass of $p=0.95$ at each decoding step (averaged over heads and layers), while horizontal lines denote fixed $k$ budgets that fail to adapt to step-wise variations in attention distribution.}
        \label{fig:topk_budget}
    \end{subfigure}

    \vspace{0.3em}

    \begin{subfigure}{\linewidth}
        \includegraphics[width=\linewidth]{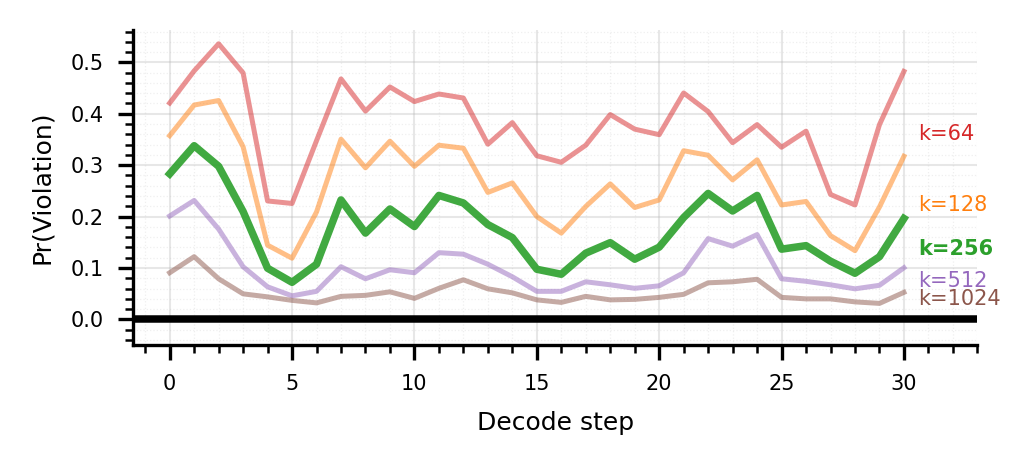}
        % \caption{Violation rates of different token budgets. The plot displays the empirical probability of an attention head violating the target mass ($p=0.95$). While adaptive top-p (black) maintains a zero-violation baseline by construction, fixed-k strategies exhibit high failure rates throughout decoding. Even $k=256$, the exact expectation of the top-p usage ($\mathbb{E}[B_{\text{adaptive}}]$) violates the constraint in over 20\% of attention heads during difficult steps, demonstrating that no single fixed budget can consistently satisfy reliability requirements for challenging retrieval tasks.}
        \caption{Violation rates across sparse attention budgets (NIAH 8K). We report the fraction of attention heads violating the target mass ($p=0.95$) during decoding. Adaptive top-p incurs zero violations, while fixed-$k$ strategies fail frequently. The expected mean budget $k=256$ even violates the constraint in over 20\% of heads, indicating that no single fixed budget reliably satisfies top-p requirements.}
        \label{fig:attn_violation}
    \end{subfigure}

    \caption{Limitations of fixed token budgets for sparse attention.}
    \label{fig:attn_budget}
\end{figure}

While simple, fixed-budget designs assume that a single budget suffices across heads, layers, and decoding steps. In practice, attention distributions vary substantially, leading to severe budget imbalance: a budget that is adequate for some heads may be overly conservative for others while remaining insufficient for more diffuse cases. Figure~\ref{fig:topk_budget} shows that the token budget required to maintain a fixed attention mass varies significantly across decoding steps. Figure~\ref{fig:attn_violation} further shows that enforcing any fixed budget, even the expected mean budget ($k=256$), results in non-trivial violations of the target mass ($p=0.95$) across layers. These results indicate that no single fixed budget can reliably satisfy top-p constraints, motivating adaptive budget allocation and highlighting the inherent limitations of top-k sparse attention.

% \textcolor{red}{Experiment 1: Top-k budget changes across heads / layers / tasks. }

% \begin{figure}[t]
%     \centering
%     \begin{subfigure}{\linewidth}
%         \includegraphics[width=\linewidth]{fig/topp_index_layer.png}
%         \caption{Mean top-p token index of layers across decoding steps. }
%         \label{fig:topp_index_layer}
%     \end{subfigure}

%     \vspace{0.3em}

%     \begin{subfigure}{\linewidth}
%         \includegraphics[width=\linewidth]{fig/topp_index_head.png}
%         \caption{Top-p token index of heads acorss decoding steps. }
%         \label{fig:topp_index_head}
%     \end{subfigure}

%     \caption{Top-p token budgets vary across attention heads, layers and decoding steps.
%     The top-p token index denotes the minimum number of tokens required to preserve a fixed cumulative attention mass \(p\).}
%     \label{fig:topp_index}
% \end{figure}

% Top-p sparse attention addresses this issue by constraining the preserved attention mass rather than the token count, by retaining the minimum set of tokens whose cumulative attention exceeds a target threshold. Figure \ref{fig:attn_budget} shows that top-p automatically adapts the compute budget and achieves a near-zero violation rate. Existing Top-p methods consists of two stages:
% (1) top-p estimation, which identifies a subset of tokens that preserves a target fraction of attention mass, and
% (2) sparse attention computation, which computes attention over the selected subset.
% Practical deployment of top-p sparse attention demands both high accuracy and low computational cost, making it an inherently multi-objective optimization problem.
% We identify that top-p sparse attention spans three interdependent objectives: top-p estimation accuracy, selection overhead, and sparse attention computation cost, shown in Table \ref{tab:top_p_comparison}. 

Top-p sparse attention addresses this issue by constraining preserved attention mass rather than token count, retaining the minimal set of tokens whose cumulative attention exceeds a target threshold. As shown in Figure~\ref{fig:attn_budget}, top-p adapts the compute budget and achieves near-zero violations. In practice, top-p sparse attention comprises two stages: (1) top-p estimation to identify tokens preserving a target attention mass, and (2) sparse attention computation over the selected subset. Efficient deployment therefore requires jointly optimizing estimation accuracy, selection overhead, and sparse attention cost. (Table~\ref{tab:top_p_comparison}).

% top-p naturally adapts to heterogeneous attention distributions and alleviates budget imbalance across heads and layers. 

% While prior work such as Twilight has demonstrated the advantages of top-p pruning, this observation highlights a key motivation for our work: effective sparse attention must not only support top-p objectives, but also account for the structural and computational challenges that arise when estimating attention mass efficiently in practice.

% \textcolor{red}{Experiment 2: Accuracy comparison between top-k and top-p of same (average) budget. }

% \subsection{Challenge: Efficient Top-P Sparse Attention}

\begin{table}[t]
\centering
\caption{Comparison of Top-$p$ accuracy and cost across methods.}
\begin{tabular}{lccc}
\hline
\multirow{2}{*}{Method} 
& \multicolumn{2}{c}{Top-$p$} 
& Sparse Attention \\
\cline{2-3}
& Accuracy & Cost & Cost \\
\hline
Token Top-p        & \textcolor{orange}{Medium} & \textcolor{red}{High} & \textcolor{MediumSeaGreen}{Low} \\
Cluster            & $\times$ & $\times$ & \textcolor{orange}{Medium} \\
\textbf{Double-P (Ours)}    & \textcolor{MediumSeaGreen}{High} & \textcolor{MediumSeaGreen}{Low} & \textcolor{MediumSeaGreen}{Low} \\
\hline
\end{tabular}
%\caption{Comparison of Top-$p$ accuracy and cost across methods.}
\label{tab:top_p_comparison}
\end{table}

Prior work shows that efficiently supporting top-p sparse attention is challenging. Token-level approaches \cite{lin2025twilight} rely on fixed token budgets for top-p estimation, which cannot reliably recover a target attention mass due to variability across heads and layers, forcing conservative budget choices. Moreover, token-level estimation incurs high selection overhead that scales linearly with context length, limiting end-to-end efficiency even with aggressive pruning.

In contrast, cluster-based sparse attention methods \cite{chen2025retroinfer} reduce sparse attention cost by operating on cluster-level representations, but both cluster selection and sparse attention computation still rely on fixed budgets, effectively reverting to top-k behavior. Consequently, these methods lack explicit control over preserved attention mass and cannot provide top-p accuracy guarantees.

To address these challenges, we propose \textbf{Double-P}, a \textbf{hierarchical top-p sparse attention framework} that reduces both top-p estimation and sparse attention cost while preserving attention mass guarantees. Double-P first performs \textbf{cluster-level Top-p attention score estimation} using size-weighted centroids, and then applies a \textbf{second-stage adaptive token budget allocation} to adaptively determine which clusters require exact token-level attention. Sparse attention is computed by combining exact attention over high-impact tokens with centroid-based approximations for the remainder. 
% By aligning probability-based pruning with the hierarchical structure of modern sparse attention systems, Double-P enables \textbf{attention value preservation} across heterogeneous heads, layers, and decoding steps. 
We further introduce \textbf{GPU efficient kernels} that minimize top-p selection overhead and maximize data locality for sparse attention computation. Extensive evaluation shows that Double-P achieves up to \textbf{2.27×} self-attention speedup and \textbf{1.26×} end-to-end decoding speedup, forming a \textbf{superior accuracy–efficiency Pareto frontier} over existing token-level and cluster-based methods (Figure~\ref{fig:latency_vs_accuracy}).

\section{Background}

\subsection{LLM and Attention Mechanism}

Large Language Models (LLMs) are built upon the Transformer architecture \cite{yang2025qwen3, liu2025deepseek, grattafiori2024llama, team2023gemini}, where the core computational primitive is the Scaled Dot-Product Attention \cite{vaswani2017attention}. In autoregressive inference, generation proceeds in two stages: a prefill stage, which processes the input prompt in parallel to construct the key–value (KV) cache, and a decoding stage, where tokens are generated sequentially using the cached representations \cite{kwon2023efficient}. During decoding stage, each output token is generated by attending over all previously seen tokens stored in the key–value (KV) cache \cite{kwon2023efficient}. Given a query token vector $q \in \mathbb{R}^d$ at a decoding step and the cached key and value matrices $K, V \in \mathbb{R}^{N \times d}$, where $d$ is the head dimension and $N$ is the current context length, the attention output $o$ is computed as:\begin{equation}o = \text{Softmax}\left(\frac{\mathbf{q} \mathbf{K}^\top}{\sqrt{d}}\right) V = \sum_{i=1}^{N} A_i v_i\end{equation}where $A_i$ represents the attention weight for the $i$-th token. In the context of long-context inference, $N$ (the sequence length) can reach hundreds of thousands. Since the computational complexity and memory bandwidth requirements scale quadratically with $N$, the full attention mechanism becomes the primary latency bottleneck.

\subsection{Sparse Attention}

To reduce the overhead of full attention, sparse attention methods approximate the output by identifying and attending only to the most critical subset of tokens, i.e., those with high attention weights $A_i$. 

\textbf{Static Selection and Permanent Eviction.} Static selection methods \cite{li2024snapkv} rely on predefined sparse attention structures, whereas permanent eviction methods \cite{xiao2023efficient, zhang2023h2o} reduce attention cost by permanently removing tokens deemed unimportant during long-context processing. 

% \textbf{Page Retrieval.} Methods such as \cite{tang2024quest, xiao2024duoattention, yang2025lserve} partition the sequence into pages and maintain coarse token statistics (e.g., key bounds) to estimate page-level relevance, selecting a subset of pages for fine-grained attention.

% \textbf{Cluster Retrieval.} Approaches including \cite{liu2025clusterkv, chen2025retroinfer, zhu2025tactic} group semantically similar tokens into clusters and compute attention scores between the query and cluster centroids to retrieve the most relevant clusters.

% \textbf{Approximate Nearest Neighbor Search.} Retrieval-based methods \cite{liu2024retrievalattention, zhang2025pqcache, chen2024magicpig} leverage approximate search structures such as RoarGraph, product quantization (PQ), or locality-sensitive hashing (LSH) to identify high-scoring tokens.

\textbf{Retrieval-Based Methods.}
Retrieval-based sparse attention methods aim to identify a subset of relevant tokens before computing exact attention. Existing approaches adopt different retrieval granularities, including page-level retrieval using coarse token statistics \cite{tang2024quest, xiao2024duoattention, yang2025lserve}, cluster-level retrieval via centroid-based similarity estimation \cite{liu2025clusterkv, chen2025retroinfer, zhu2025tactic, liu2025juno++}, and token-level approximate nearest neighbor search using structures such as product quantization, locality-sensitive hashing, or graph-based indices \cite{liu2024retrievalattention, zhang2025pqcache, chen2024magicpig}. While these methods significantly reduce attention cost, their retrieval quality is often sensitive to approximation error and fixed selection budgets.

\textbf{Prefill Acceleration.} Prefill acceleration methods \cite{jiang2024minference, gao2024seerattention, zhang2025spargeattn, xu2025xattention, dalton2026quoka} focus on reducing the cost of prompt processing by exploiting sparsity in attention computation.

\textbf{Trainable Sparse Attention.} Several works \cite{yuan2025native, lu2025moba, liu2025deepseek} explore trainable sparse attention by learning content-based routing or token selection mechanisms.

% While effectively reducing computation, the retrieved-based sparse attention methods all rely on a \textbf{Top-k} selection policy, where a fixed budget of tokens is retrieved regardless of the actual attention distribution.

% \textcolor{red}{[wt: Revise the following section introducing Twilight. ]}

\textbf{Top-P Sparse Attention.} Recent work \cite{lin2025twilight} extends sparse attention from fixed top-$k$ to probability-based top-$p$ selection, adaptively retaining tokens whose cumulative attention mass exceeds a target threshold. 

% To overcome the limitations of fixed top-k budget algorithms, recent work such as \textbf{Twilight} \cite{lin2025twilight} introduces a \emph{select-then-prune} pipeline to enable \textbf{top-p} (probability-based) sparse attention. The goal of top-p attention is to retain a variable number of tokens whose cumulative attention mass exceeds a target threshold $p$, providing an explicit accuracy guarantee that adapts to the underlying attention distribution.

% Because exact top-p selection requires computing and sorting attention scores over all tokens, Twilight approximates this process through token-level estimation and pruning. It first identifies a candidate set using low-cost approximate scores, refines attention estimates within this set, and then applies top-p pruning to select tokens whose estimated probability mass meets the target threshold, before computing exact attention on the retained tokens.

\subsection{Other Inference Optimizations}

\textbf{Layer-wise and head-wise budget allocation.} These methods \cite{feng2024ada, tang2024razorattention, xiao2024duoattention} allocate KV cache budgets unevenly across layers or attention heads based on their estimated importance, reducing memory and computation while preserving model quality.

\textbf{Quantization.} 
Prior work \cite{frantar2022gptq, lin2024awq, tseng2024quip, van2024gptvq, liu2025vq} investigates low-precision quantization for large language models to reduce memory footprint and inference cost. In addition to weight quantization, recent methods \cite{hooper2024kvquant, yue2024wkvquant, kang2024gear} further compress the KV cache using reduced precision, achieving substantial memory savings with minimal accuracy degradation. 

\textbf{Linear Attention.} Linear attention methods \cite{katharopoulos2020transformers, gu2024mamba} replace softmax attention with kernelized or state-space formulations to achieve linear-time sequence processing and eliminate explicit KV cache storage.

\section{Motivation}

Top-p sparse attention offers an appealing alternative to fixed-budget top-k methods by directly constraining preserved attention mass and providing explicit accuracy guarantees. However, realizing these benefits efficiently in practice remains challenging. 

To overcome the limitations of fixed top-k budget algorithms, recent work such as Twilight \cite{lin2025twilight} introduces probability-based top-$p$ sparse attention, which adaptively retains a variable number of tokens whose cumulative attention mass exceeds a target threshold $p$. Since exact top-p selection requires computing and sorting attention scores over all tokens, Twilight adopts a \emph{select-then-prune} pipeline that relies on approximate token-level attention estimation to identify a candidate set, applies top-p pruning on the estimated scores, and computes exact attention only on the retained tokens.

However, token-level top-p methods reveal two fundamental limitations. First, top-p estimation is typically performed under a fixed token budget, which cannot reliably recover a target attention mass due to substantial variation in attention distributions across heads, layers, and decoding steps. 

\begin{figure}[t]
    \centering
        \includegraphics[width=\linewidth]{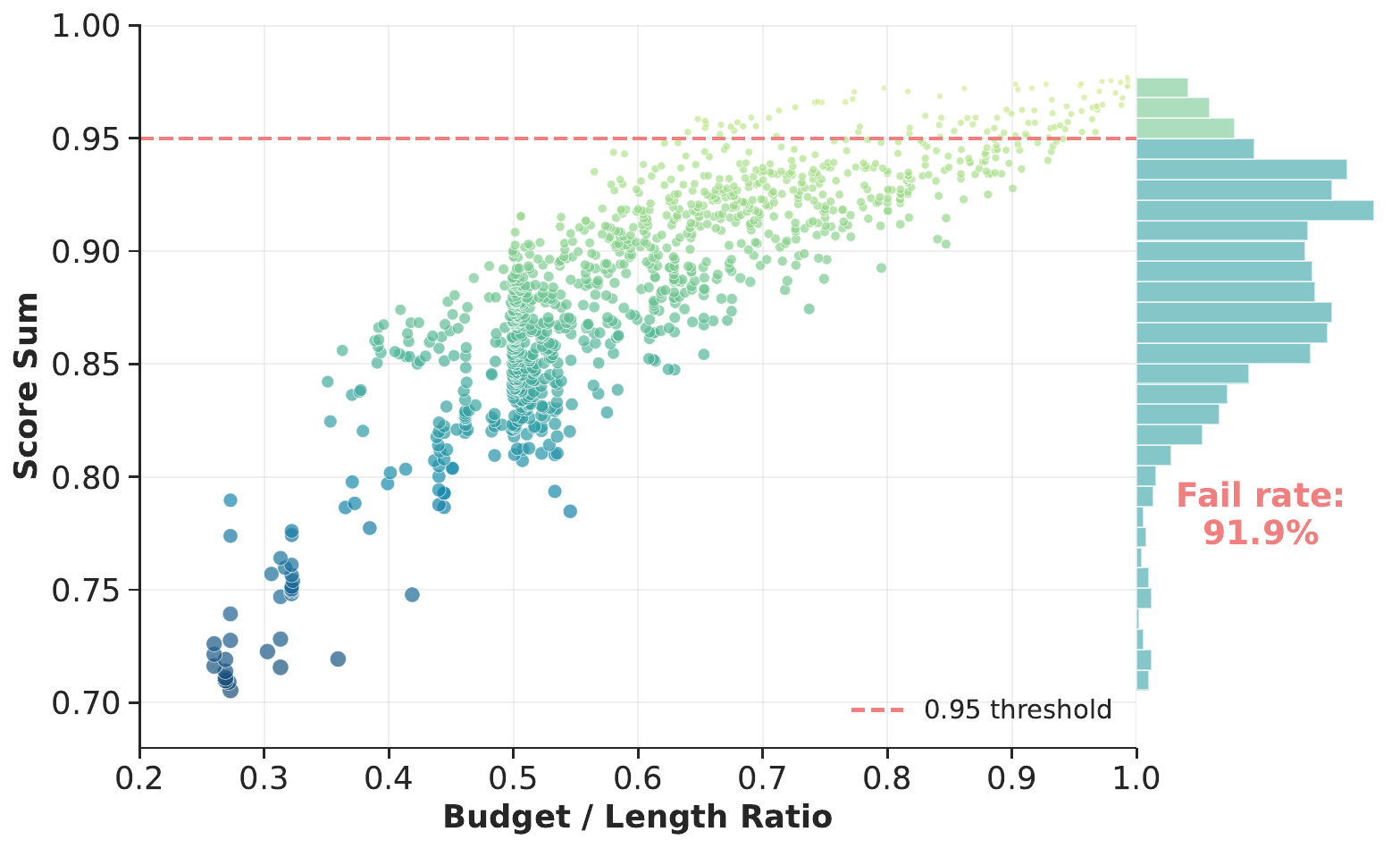}
        \caption{Failure of fixed-budget token-level Top-P estimation. Each point shows the recovered attention mass and its budget ratio for an individual LongBench sample. Under the recommended fixed budget, 91.9\% of samples fail to recover sufficient attention mass.}
        \label{fig:score_vs_budget}
\end{figure}

Figure \ref{fig:score_vs_budget} provides empirical evidence that fixed-budget token-level top-p estimation is fundamentally unreliable. Each point represents a LongBench sample, plotting the recovered attention mass against the token budget normalized by context length. Under the commonly recommended fixed budget (e.g., 8192 tokens for Quest), 91.9\% of samples fail to reach the target attention mass $p=0.95$. Moreover, the figure shows substantial variability across samples: the same budget can yield near-complete attention recovery in some cases while severely underestimating attention mass in others. This reflects highly non-uniform attention distributions across inputs and demonstrates that no single fixed token budget can reliably satisfy top-p constraints. Consequently, token-level top-p methods must adopt conservative, over-provisioned budgets to avoid failure, which directly hurts their efficiency advantages.

\begin{figure}[h]
    \centering
    \includegraphics[width=1.0\linewidth]{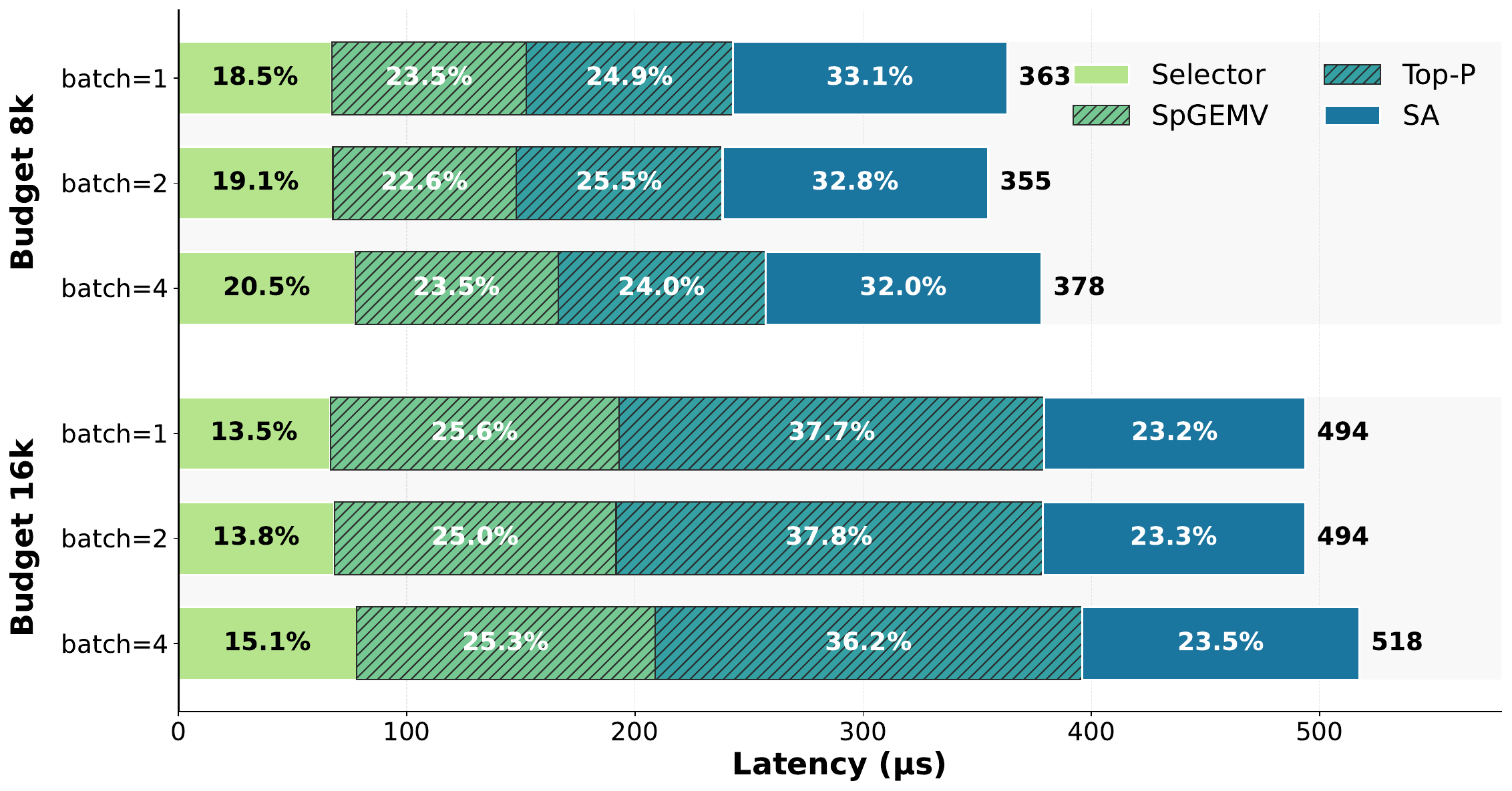}
    \caption{Latency Breakdown of Token-Level Top-p Sparse Attention. Shaded bars indicate Top-p estimation overhead. Token-level attention score estimation (SpGEMV) and Top-p selection account for a substantial fraction of total latency, indicating that top-p estimation overhead dominates performance even before sparse attention (SA) is applied.}
    \label{fig:twilight_latency_breakdown}
\end{figure}

Second, because estimation operates at the token level, the cost of top-p selection scales linearly with the number of tokens considered. 
% As a result, estimation overhead remains a dominant factor in end-to-end latency, even when aggressive pruning is applied afterward.
Figure \ref{fig:twilight_latency_breakdown} shows that attention score estimation via SpGEMV and subsequent top-p selection together account for a large fraction of the total decoding time. This overhead persists regardless of how aggressively tokens are pruned afterward, since both stages operate at the token level and scale linearly with context length. As a result, even though Twilight reduces the cost of the final sparse attention computation, the high estimation overhead fundamentally limits end-to-end speedups. This observation motivates reducing the cost of top-p estimation itself, rather than relying solely on pruning after selection.

On the other hand, cluster-based sparse attention methods \cite{chen2025retroinfer} reduce sparse attention cost by operating on cluster-level representations. While efficient, both cluster selection and token-level computation in these methods rely on fixed cluster or token budgets. Consequently, they lack explicit control over preserved attention mass and cannot provide accuracy guarantees aligned with top-p objectives.

Together, these limitations expose a key gap in the design space: existing methods either support top-p selection at high estimation cost or achieve efficiency by reverting to fixed-budget top-k behavior. Bridging this gap requires rethinking top-p sparse attention as an end-to-end problem that jointly considers estimation, selection, and sparse computation, motivating a hierarchical approach that aligns probability-based pruning with the structure of sparse attention systems.

\section{Double-P Sparse Attention}

\begin{figure*}
    \centering
    \includegraphics[width=1.0\linewidth]{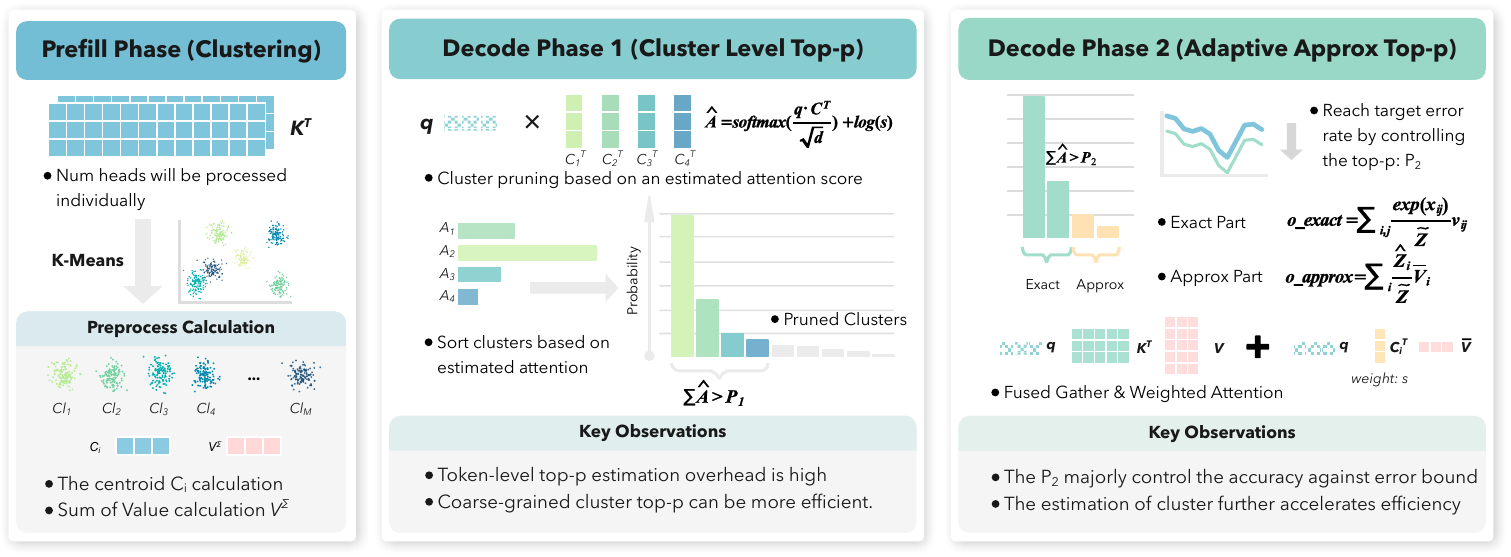}
    \caption{\textbf{Double-P framework overview.} Double-P performs hierarchical top-p sparse attention by first estimating attention mass at the cluster level using size-weighted centroids, then adaptively refining token-level computation through a second top-p stage. Exact token attention and centroid-based approximations are combined to achieve high accuracy with low estimation and sparse attention cost.}
    \label{fig:doublep-design}
\end{figure*}

In this section, we present Double-P, a hierarchical top-p sparse attention framework designed to jointly optimize top-p estimation accuracy, selection overhead, and sparse attention cost, shown in Figure \ref{fig:doublep-design}. Double-P exploits the hierarchical structure of modern sparse attention systems by first estimating attention mass at the cluster level and then adaptively refining token-level computation through a second top-p stage. This design enables accurate, probability-driven pruning while avoiding fixed token budgets.

\subsection{Prefilling}
\label{sec:cluster_metadata}

During the prefill stage, the input context is processed once to construct the key–value (KV) cache, where keys and values are stored as matrices $\mathbf{K}, \mathbf{V} \in \mathbb{R}^{N \times d}$, with $N$ denoting the context length and $d$ the attention head dimension. To enable coarse-grained attention estimation, we partition the KV cache into clusters by applying k-means clustering \cite{hartigan1979algorithm} over the key vectors $\mathbf{K}$. This groups tokens with similar key representations into semantically coherent clusters. 

% we exploit its structural organization by partitioning tokens into fixed-size pages or semantic clusters. This follows layouts already adopted by many sparse attention and KV management systems \cite{tang2024quest, liu2025clusterkv, chen2025retroinfer}. Each cluster groups a contiguous or semantically related set of tokens and serves as the basic unit for coarse-grained attention estimation during subsequent decoding.

For each cluster \(i\) with token set \(\mathcal{T}_i\), cluster centroid $\mathbf{C}_i$ size \(s_i = |\mathcal{T}_i|\), we aggregate token values into a single cluster-level representation by summing over tokens,
\begin{equation}
\mathbf{V}^{\Sigma}_i = \sum_{j\in \mathcal{T}_i}\mathbf{v}_{ij},
\qquad
\bar{\mathbf{V}}_i = \frac{1}{s_i}\mathbf{V}^{\Sigma}_i,
\end{equation}
where $\mathbf{v}_{ij}$ is the $j$-th value vector in $i$-th cluster, \(\mathbf{V}^{\Sigma}_i\) denotes the sum of values and \(\bar{\mathbf{V}}_i\) its normalized average.

Together, the cluster centroid \(\mathbf{C}_i\), size \(s_i\), and value aggregates provide a compact and efficient summary of the KV cache. These representations enable low-cost estimation of attention mass at the cluster level and support mixed-granularity attention computation in later stages of Double-P.

\subsection{Stage 1: Cluster-Level Attention Score Estimation}

% First Top-P: Approximate attention mass. Perform Top-P over clusters. 

% \begin{equation}
%     \hat{A} = \text{softmax} (\frac {\mathbf{q} \cdot \mathbf{C_i}} {\sqrt d} + \log s_i)
% \end{equation}

% Select ~50\% clusters to recover target attention mass. Still too expensive to compute token-level attention on all of them

% \subsection{Cluster / Page Centroids}
% \label{sec:cluster_centroids}

% During the prefill stage, when the full input context is processed and the KV cache is constructed, we exploit its structural organization by partitioning tokens into fixed-size pages or semantic clusters, following the layouts already adopted by many sparse attention and KV management systems \cite{tang2024quest, liu2025clusterkv, chen2025retroinfer}. Each cluster groups a contiguous or semantically related set of tokens and serves as the basic unit for coarse-grained attention estimation in subsequent decoding.

% For each cluster $i$, we maintain two pieces of metadata:
% (1) a \emph{cluster centroid} $\mathbf{C}_i$, computed as the average of the key vectors within the cluster, and
% (2) the \emph{cluster size} $s_i$, defined as the number of tokens contained in the cluster.
% Together, these quantities provide a compact summary of both the semantic content and scale of the cluster, which we leverage for efficient attention estimation.

During decoding stage, given a query vector $\mathbf{q} \in \mathbb{R}^{1 \times d}$, consider a cluster $i$ of tokens $\mathcal{T}_i$ with centroid $\mathbf{C}_i \in \mathbb{R}^{1 \times d}$ and size $s_i$. We first define the corresponding attention logits as 
\begin{equation}
x_{ij} = \frac{\mathbf{q} \mathbf{k}_{ij}^\top}{\sqrt{d}},
\qquad
\bar{x}_i = \frac{\mathbf{q} \mathbf{C}_i^\top}{\sqrt{d}}.
\end{equation}
where $\mathbf{k}_{ij}$ is the $j$-th key vector in $i$-th cluster, $x_{ij}$ denotes the exact token-level attention logit and $\bar{x}_i$ is its centroid-based approximation. 
The exact cluster mass is 
\begin{equation}
    Z_i = \sum_{j\in \mathcal{T}_i}\exp(x_{ij}). 
\end{equation}
Computing $Z_i$ exactly requires evaluating all token-level logits, which is expensive. Instead, we approximate the cluster mass using the centroid logit scaled by the cluster size:
\begin{equation}
\widehat{Z}_i = s_i \exp(\bar{x}_i),
\qquad
\log \widehat{Z}_i = \bar{x}_i + \log s_i.
\end{equation}
Let $\mathbf{C} \in \mathbb{R}^{K \times d}$ denote the matrix of all cluster centroids, with rows $\mathbf{C}_i$, and let $s \in \mathbb{R}^K$ be the vector of cluster sizes. 
Normalizing these approximate masses yields an estimated cluster-level attention distribution, 
\begin{equation}
\widehat{\mathbf{A}}
=
\frac{\widehat{\mathbf{Z}}}{\sum_{\ell}\widehat{Z}_{\ell}}
=
\mathrm{softmax}\!\left(\frac{\mathbf{q} \mathbf{C}^\top}{\sqrt{d}} + \log \mathbf{s}\right),
\end{equation}
where $\widehat{\mathbf{A}}$ serves as an approximation of the attention mass assigned to the clusters. 

% Given a query vector $\mathbf{q}$, we estimate the attention contribution of each cluster using a size-weighted centroid score. Specifically, we compute
% \begin{equation}
% A_i = \mathrm{softmax}\!\left(
% \frac{\mathbf{q} \cdot \mathbf{C}_i}{\sqrt{d}} + \log s_i
% \right),
% \end{equation}
% where $d$ denotes the attention head dimension. The dot product between the query and the cluster centroid captures the average relevance of tokens within the cluster, while the logarithmic size term accounts for the multiplicity of tokens contributing to that relevance.

Compared to token-level estimation, this formulation provides a low-cost yet informative estimate of how attention is distributed across clusters. We then apply a Top-P selection over clusters to identify the minimal set whose cumulative attention mass exceeds a target threshold \(p_1\)

\begin{equation}
\mathcal{C}_{p}
=
\left\{
i \;\middle|\;
\sum_{i \in \mathcal{C}_{p}} \widehat{A}_i \;\ge\; p
\right\},
\end{equation}
with clusters sorted by $\widehat{A}_i$ in descending order. 

\subsection{Stage 2: Adaptive Token Budget Allocation for Sparse Attention}

\begin{figure}[t]
    \centering
    \includegraphics[width=\linewidth]{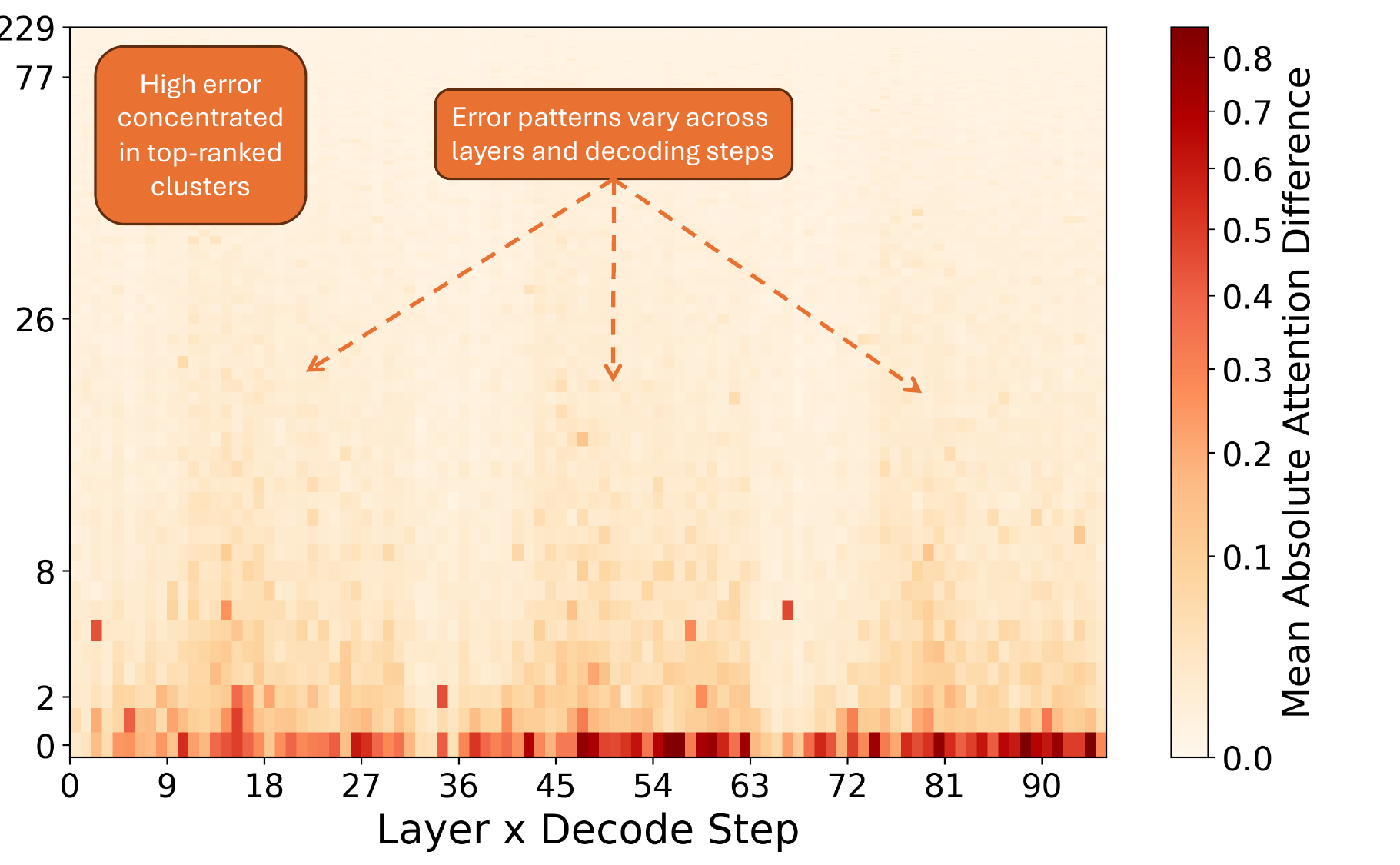}
    \caption{Heatmap of absolute error between exact token-level attention and cluster-centroid–based approximation, indexed by cluster rank. High approximation error is concentrated in a small number of top-ranked clusters, while the error distribution varies significantly across layers and decoding steps.}
    \label{fig:diff_heatmap}
\end{figure}

While the first-stage cluster-level Top-P provides a coarse estimate of attention mass, computing exact token-level attention for all selected clusters remains unnecessary and inefficient. Figure \ref{fig:diff_heatmap} visualizes the absolute attention error introduced by replacing token-level attention with cluster-centroid estimates across layers and decoding steps. The error is strongly long-tailed: a small number of clusters contribute the majority of approximation error, while most clusters have negligible impact. Besides, the error distribution varies significantly across layers and decoding steps. 

\begin{figure}[h]
    \centering
    \includegraphics[width=\linewidth]{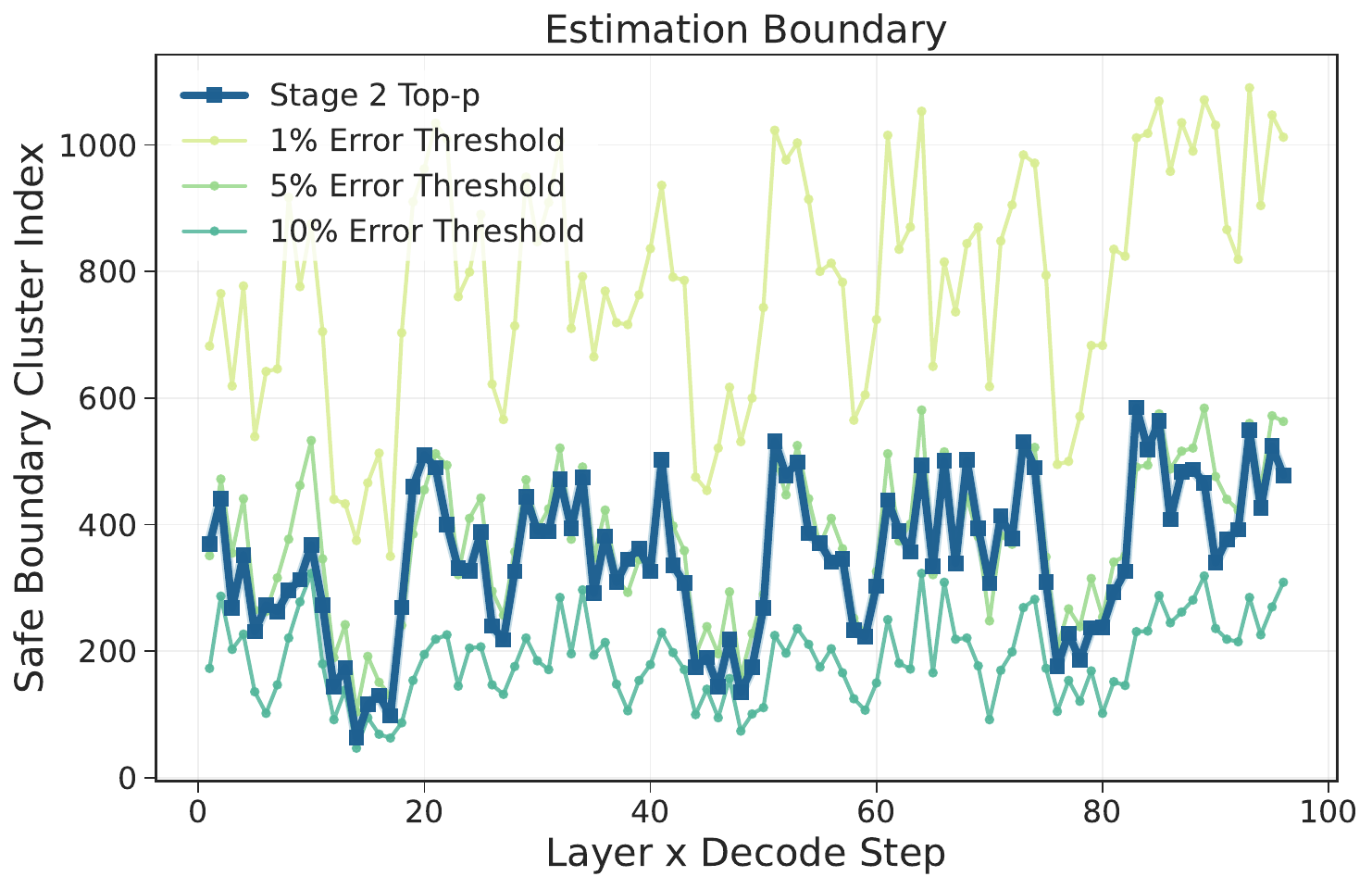}
    \caption{\textbf{Adaptive cluster selection for bounded attention error.}
    The minimum number of clusters required to satisfy a given error threshold varies significantly across layers and decoding steps. The \textbf{bold} line shows the number of clusters selected by the second-stage Top-P in Double-P, which closely tracks the minimum required clusters, demonstrating near-optimal adaptive allocation.}
    \label{fig:estimation_boundary}
\end{figure}

Figure \ref{fig:estimation_boundary} further quantifies this observation by showing the minimum number of clusters that must be computed with exact attention to satisfy a fixed error bound. The required number fluctuates widely across layers and decoding steps, indicating that enforcing a fixed cluster or token budget would either waste computation or violate accuracy guarantees.

To address this challenge, we introduce an \emph{adaptive token budget allocation} mechanism for sparse attention based on a second Top-P stage. Rather than assigning a fixed token budget, this stage refines computation by allocating token-level attention in proportion to each cluster’s estimated contribution to the overall attention mass. Clusters with higher impact are assigned exact token-level computation, while clusters with lower impact are approximated using centroid-based summaries. The emphasized curve shown in Figure \ref{fig:estimation_boundary} corresponds to the clusters selected by the second-stage top-p in Double-P, which closely matches the minimum required clusters given a certain error threshold. This result shows that the second top-p effectively allocates exact computation only where necessary, achieving near-minimal cost while maintaining bounded attention error.

After the second top-p selection, we partition the clusters into two disjoint sets: clusters that receive exact token-level attention and clusters whose contributions are approximated. Let $\mathcal{C}_{\mathrm{exact}}$ denote the set of clusters assigned to exact computation, and $\mathcal{C}_{\mathrm{approx}}$ denote the remaining clusters. 

To combine exact and approximate contributions in a unified manner, we normalize both using a shared attention mass, 
\begin{equation}
\widetilde{Z}
=
\sum_{i\in \mathcal{C}_{\mathrm{exact}}}\sum_{j\in \mathcal{T}_i}\exp(x_{ij})
+
\sum_{i\in \mathcal{C}_{\mathrm{approx}}}\widehat{Z}_i.
\end{equation}
The final attention output is then computed as a mixture of exact and approximate contributions,
\begin{equation}
\mathbf{o}(\mathbf{q})
=
\sum_{i\in \mathcal{C}_{\mathrm{exact}}}\sum_{j\in \mathcal{T}_i}
\frac{\exp(x_{ij})}{\widetilde{Z}}\,\mathbf{v}_{ij}
\;+\;
\sum_{i\in \mathcal{C}_{\mathrm{approx}}}
\frac{\widehat{Z}_i}{\widetilde{Z}}\,\bar{\mathbf{V}}_i,
\label{eq:mix_exact_approx_normalized}
\end{equation}
where exact tokens contribute through their full attention weights, and approximate clusters contribute proportionally to their estimated attention mass.

% By decoupling sparse attention cost from fixed token budgets and aligning token-level computation with estimated attention mass, the adaptive allocation stage completes the hierarchical top-p pipeline. Together with cluster-level estimation, it enables Double-P to jointly optimize estimation accuracy, selection overhead, and sparse attention cost, achieving an efficient and robust top-p sparse attention mechanism. 

\subsection{Efficient kernel implementation}

% Fast Top-P kernel on sorted tensors. 
% Fused gathering tokens and centroids into contiguous buffers. 
% Weighted Flash Attention. 

We design efficient GPU kernel implementations that minimizes selection overhead and maximizes data locality during sparse attention computation. 

\textbf{Efficient Top-p.} Both stages of Double-P rely on Top-P selection over attention scores that are already sorted. We implement a custom Top-P kernel that operates directly on sorted tensors, performing prefix-sum accumulation and early stopping to identify the minimal set of elements whose cumulative probability exceeds the target threshold. 
% (Introduce why better than twilight top-p?)

\textbf{Efficient Token \& Cluster Gathering.} After second Top-P selection, we fuse gathering selected tokens and cluster centroids into one kernel, and copy them into a contiguous memory buffer. This fusion step enables coalesced reads for subsequent attention computation.

\textbf{Efficient Sparse Attention.} We compute sparse attention using a weighted variant of FlashAttention over the fused buffers, introduced by RetroInfer \cite{chen2025retroinfer}. Exact tokens and approximate cluster summaries are processed uniformly within the same attention kernel. This design avoids separate kernels for exact and approximate attention.

\begin{figure*}[t]
    \centering
    \includegraphics[width=\linewidth]{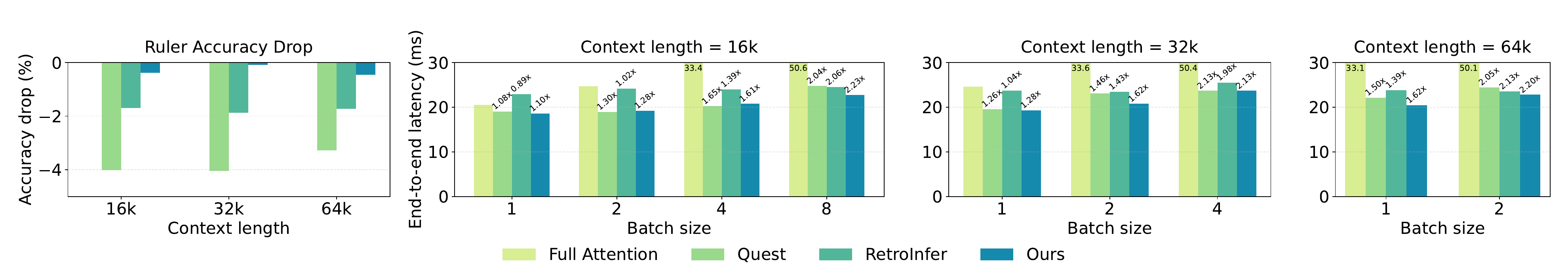}
    \caption{\textbf{Accuracy drop and end-to-end decoding latency on RULER.} Double-P achieves \textbf{near-zero} accuracy drop compared with full attention while consistently delivering faster inference.}
    \label{fig:latency_e2e}

\end{figure*}

\section{Evaluation}

We evaluate Double-P from both accuracy and efficiency perspectives to assess its effectiveness for long-context inference. Our experiments focus on (i) downstream task accuracy, and (ii) end-to-end decoding latency with detailed breakdowns of estimation and sparse attention cost. We compare against state-of-the-art baselines across multiple benchmarks, and context lengths to demonstrate robustness and practical accuracy efficiency gains.

%\subsection{Evaluation Setup}
\subsection{Experimental Setup}
\label{sec:eval_setup}

\textbf{Hardware and system configuration.}
All experiments are conducted on a single-node GPU server equipped with one NVIDIA H100 PCIe GPU (80GB memory) and an Intel(R) Xeon(R) Gold 5416S CPU, using CUDA 12.8 and PyTorch 2.8. 

\textbf{Models.}
We evaluate Double-P on the representative open-source long-context LLM: LLaMA3.1-8B \cite{grattafiori2024llama}, with context lengths up to 128K tokens. It has 32 layers and 32 heads, adopts group query attention (GQA) with group size 4.

\textbf{Baselines.}
We compare Double-P against representative sparse attention baselines covering both page-level and cluster-based designs. For fair comparison, all methods preserve sink tokens (4 tokens) and sliding-window tokens (64 tokens) at every layer \cite{xiao2023efficient}, while sparse attention is applied only to the remaining middle tokens.
(1) \textbf{Quest} \cite{tang2024quest} is a page-level method that selects top-k pages based on key bounds; we follow the default page size 16 and retrieve 25\% of the tokens.
(2) \textbf{RetroInfer} \cite{chen2025retroinfer} is a cluster-based method that retrieves a fixed number of top-k clusters and performs sparse attention over tokens and cluster centroids; we use the cluster configurations and retrieval budgets suggested by the authors. 
(3) \textbf{Twilight} \cite{lin2025twilight} is a token-level top-p sparse attention method that performs select-then-prune using approximate token-level attention estimation under a fixed selection budget; we use the recommended 25\% token budget from the original paper on Quest.

\textbf{Benchmarks.}
We evaluate accuracy on two established long-context benchmarks.
(1) \textbf{RULER} \cite{hsieh2024ruler} is a comprehensive benchmark consisting of 13 tasks, with context lengths ranging from 4K to 128K tokens. 
(2) \textbf{LongBench} \cite{bai2024longbench} focuses on realistic long-context applications and comprises six major task categories with 21 individual tasks. The average input length of most tasks ranges from 5K to 15K tokens.

% Models: Llama 3.1 8b, Qwen 3 8b. 

% Benchmarks: Ruler, LongBench. 

% Baselines: Quest, Quest + Twilight, RetroInfer, RetroInfer + Twilight. 

\subsection{Accuracy}

\begin{table}[h]
\centering
\small
\setlength{\tabcolsep}{3pt}
\caption{Accuracy comparison on Ruler and LongBench for LLaMA3.1-8B.}
\renewcommand{\arraystretch}{1.1}
\begin{tabular}{lccccccc}
\hline
Method 
& 16k 
& 32k 
& 64k 
& \textbf{Avg.}
& LongBench \\
\hline

LLaMA-3.1-8B             & 93.25 & 90.00 & 85.36 & 89.54 & 39.33 \\
Quest                    & 89.24 & 85.95 & 83.61 & 86.27 & 38.76 \\
RetroInfer               & 91.56 & 88.12 & 83.77 & 87.82 & 39.03 \\
Quest + Twi              & 86.73 & 86.50 & 81.87 & 85.03 & 38.97 \\
% RetroInfer + Twi         & 92.83 & \textbf{89.95} & \textbf{85.21} & \textbf{89.33} &  \\
\textbf{Double-P (Ours)} & \textbf{92.87} & \textbf{89.91} & \textbf{84.55} & \textbf{89.08} & \textbf{39.06} \\
                         & \textbf{\textcolor{DarkGreen}{+1.31}} & \textbf{\textcolor{DarkGreen}{+1.79}} & \textbf{\textcolor{DarkGreen}{+0.78}} & \textbf{\textcolor{DarkGreen}{+1.26}} & \textbf{\textcolor{DarkGreen}{+0.03}} \\
\hline

\end{tabular}
\label{tab:accuracy_llama}
\end{table}

% \begin{table}[h]
% \centering
% \small
% \renewcommand{\arraystretch}{1.1}
% \caption{LongBench average scores. }
% \begin{tabular}{lcc}
% \hline
% Method & LLaMA-3.1-8B & Qwen3-8B \\
% \hline
% Full Attention        & 39.33 & 35.82 \\
% Quest                 & 38.76 & 35.52 \\
% RetroInfer            & 39.03 & 35.63 \\
% Quest + Twi           & 38.97 & 35.26 \\
% % RetroInfer + Twi      &  &  \\
% \textbf{Double-P (Ours)}     & \textbf{39.06} & \textbf{35.78} \\
% \hline
% \end{tabular}
% \label{tab:longbench}
% \end{table}

\textbf{Accuracy on RULER. }
Table \ref{tab:accuracy_llama} reports average RULER accuracy for LLaMA-3.1-8B across context lengths. Double-P consistently achieves the highest accuracy among sparse attention methods, improving over the strongest baseline by \textbf{+1.31}, \textbf{+1.79}, and \textbf{+0.78} absolute points at 16K, 32K, and 64K context lengths, respectively. On average, Double-P yields a \textbf{+1.26} point gain and achieves \textbf{near-zero accuracy drop} compared with full attention. These results demonstrate that Double-P preserves attention mass more reliably than prior methods, translating directly into higher accuracy under long-context inference.

\textbf{Accuracy on LongBench. }
Table \ref{tab:accuracy_llama} reports performance on LongBench. Double-P achieves the highest accuracy among all sparse attention methods and remains within a near-zero gap to full attention. Double-P maintains strong generalization across heterogeneous LongBench tasks, demonstrating that its probability-driven selection strategy preserves attention quality beyond synthetic benchmarks.

We report the full per-task accuracy results in the appendix.

% \textcolor{red}{Be more quantitative to show the strengths. }

% \textcolor{red}{Table 1. Results on LongBench.}

% \textcolor{red}{Table 2. Results on Ruler.}

\subsection{Efficiency}

\begin{figure}[t]
    \centering
    \includegraphics[width=1.0\linewidth]{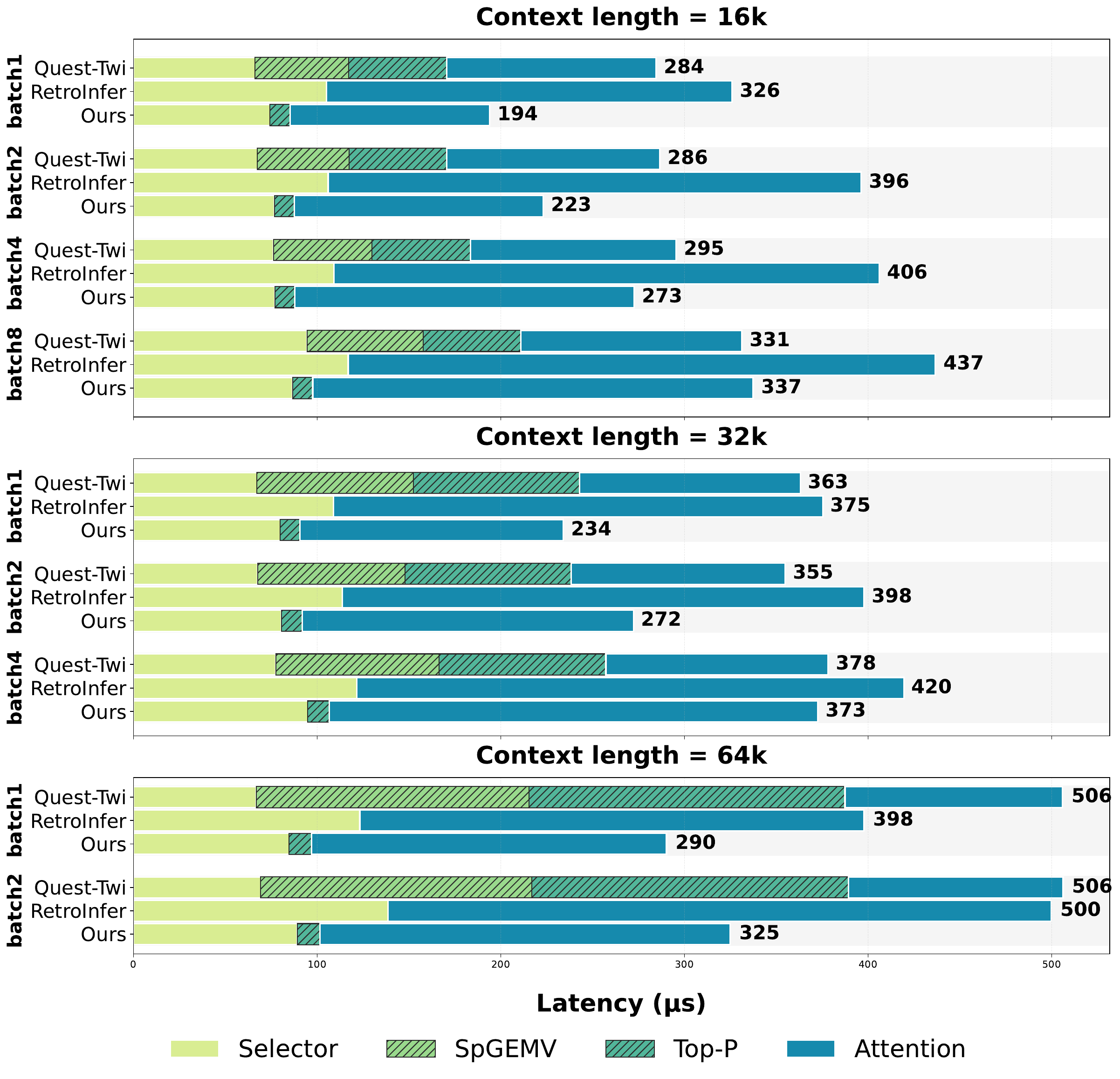}
    \caption{\textbf{Attention latency breakdown.} Shaded bars indicate top-p estimation overhead. Double-P minimizes this cost and achieves significant attention-level speedups compared to prior sparse attention methods.}
    \label{fig:latency_breakdown}
\end{figure}

We evaluate the efficiency of Double-P and other baselines with different context lengths and batch sizes. We use Llama-3.1-8B and NIAH to measure both attention latency breakdown and end-to-end decoding latency. 

\textbf{Attention latency.}
Figure \ref{fig:latency_breakdown} shows the breakdown of attention computation across sparse attention methods, where \textbf{shaded bars indicate top-p estimation overhead}. In existing top-p approaches, token-level estimation dominates attention latency, accounting for \textbf{over 60\%} of the total cost. Double-P largely eliminates this overhead by replacing token-level estimation with efficient cluster-level approximation. In addition, its \textbf{adaptive token budget allocation} significantly reduces the cost of the final sparse attention step by avoiding unnecessary token-level computation on low-impact clusters. Together, these effects enable Double-P to achieve up to \textbf{1.74$\times$} and \textbf{1.78$\times$} attention-level speedup over Quest-Twilight and RetroInfer, respectively.

\textbf{End-to-end decoding latency.}
Figure \ref{fig:latency_e2e} reports the end-to-end decoding latency per output token across sparse attention methods, together with the corresponding accuracy drop relative to full attention. Double-P achieves up to \textbf{1.11$\times$} and \textbf{1.26$\times$} speedup over Quest and RetroInfer, respectively, while maintaining a \textbf{near-zero accuracy drop} compared with full attention. Moreover, Double-P outperforms full attention by \textbf{up to 2.23×}, demonstrating that its algorithmic efficiency translates directly into practical decoding speedups without sacrificing accuracy.

% \textcolor{red}{Figure 1. End-to-end Latency. Comparing Full Attention, Quest + Twilight, RetroInfer + Twilight, Ours. On diferent batch sizes and context sequence lengths. }

% \textcolor{red}{Figure 2. Latency breakdown. Comparing Twilight. }

\subsection{Ablation Study}

% \begin{figure*}
%     \centering
%     \includegraphics[width=0.9\linewidth]{fig/ablation_tradeoff.pdf}
%     \caption{Caption}
%     \label{fig:ablation}
% \end{figure*}

% \begin{figure}[t]
%     \centering
%     \begin{subfigure}{\linewidth}
%         \includegraphics[width=0.95\linewidth]{fig/ablation_tradeoff_32k_cwe.pdf}
%         \label{fig:ablation_cwe}
%     \end{subfigure}

%     \vspace{0.3em}

%     \begin{subfigure}{\linewidth}
%         \includegraphics[width=0.95\linewidth]{fig/ablation_tradeoff_32k_fwe.pdf}
%         \label{fig:ablation_fwe}
%     \end{subfigure}

%     \caption{Accuracy–efficiency trade-off under different Top-P configurations.}
%     \label{fig:ablation}
% \end{figure}

\begin{figure}
    \centering
    \includegraphics[width=1.0\linewidth]{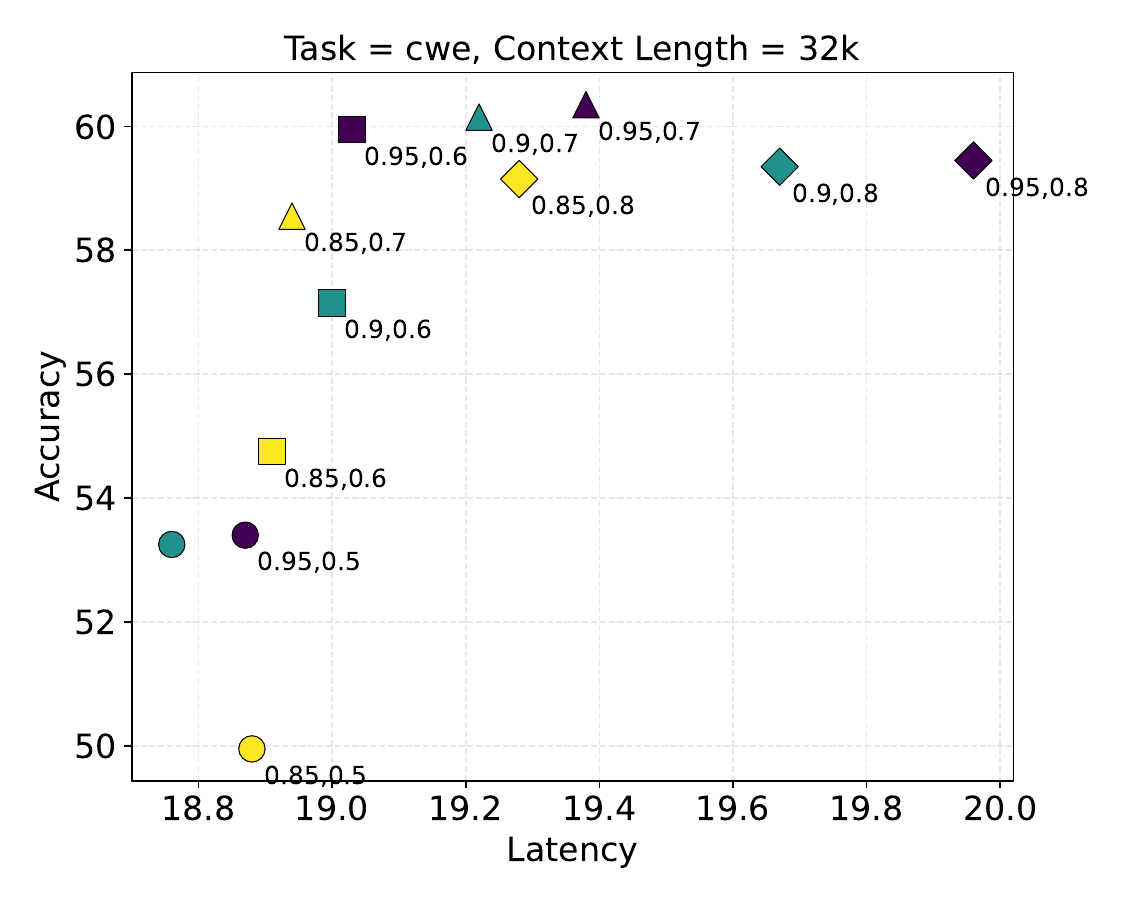}
    \caption{Accuracy and latency of different Double-P configurations tested on Ruler 32k cwe task.}
    \label{fig:ablation}
\end{figure}

% The accuracy and cost of two top-ps. (Attention score estimation accuracy, and cost). 

Figure \ref{fig:ablation} analyzes the sensitivity of Double-P to the two top-$p$ thresholds controlling attention mass preservation at the cluster level ($p_1$) and token-level refinement ($p_2$). Varying these parameters reveals a clear accuracy–efficiency trade-off: larger $p_1$ and $p_2$ retain a greater fraction of attention mass and improve accuracy, while smaller values reduce computation and yield higher efficiency. In practice, we select $(p_1, p_2) = (0.95, 0.7)$ for LLaMA-3.1-8B, which provides a balanced trade-off between accuracy and speed.

We further evaluate Double-P on the Qwen-3-8B model \cite{yang2025qwen3}, with $(p_1, p_2) = (0.99, 0.8)$. The corresponding accuracy results on RULER and LongBench are reported in Table \ref{tab:accuracy_qwen}.

\begin{table}[h]
\centering
\small
\caption{Accuracy comparison on Ruler and LongBench for Qwen3-8B.}
\setlength{\tabcolsep}{3pt}
\renewcommand{\arraystretch}{1.1}
\begin{tabular}{lccccccc}
\hline
Method 
& 16k 
& 32k 
& 64k 
& \textbf{Avg.} 
& LongBench \\
\hline

Qwen3-8B                 & 91.25 & 90.59 & 80.40 & 87.41 & 35.82 \\
Quest                    & 77.12 & 79.80 & 60.12 & 72.35 & 35.52 \\
RetroInfer               & 91.16 & 90.65 & 80.09 & 87.3 & 35.63 \\
Quest + Twi              & 76.69 & 79.50 & 58.59 & 71.59 & 35.26 \\
\textbf{Double-P (Ours)} & 91.02 & 90.36 & 79.26 & 86.88 & 35.78 \\
\hline
\end{tabular}
\label{tab:accuracy_qwen}
\end{table}

\section{Conclusion}

This paper proposes \textbf{Double-P}, a hierarchical top-p sparse attention framework for efficient long-context LLM inference. By jointly optimizing attention mass estimation, candidate selection, and sparse attention, Double-P aligns probability-based pruning with the hierarchical structure of sparse attention systems, eliminating fixed-budget constraints and enabling adaptive allocation of token-level computation across decoding steps. As a result, Double-P preserves top-p attention guarantees and consistently achieves higher accuracy than prior sparse attention methods, remaining within a near-zero gap to full attention. At the same time, Double-P delivers substantial efficiency improvements, achieving up to $1.78\times$ attention-level speedup over token-level top-$p$ methods and up to $1.26\times$ end-to-end decoding speedup over strong sparse attention baselines, while accelerating inference by up to $2.23\times$ compared to full attention. These findings demonstrate that effective top-p sparse attention requires adaptive, probability-driven computation across multiple levels of abstraction, and position Double-P as a principled and scalable foundation for future sparse attention designs in long-context language models.

\newpage

\section*{Acknowledgements}

This paper is supported in part by SRC JUMP 2.0 Center for Processing with Intelligent Storage and Memory (PRISM) and Center for Machine-Integrated Computing and Security (MICS).

% \textbf{Do not} include acknowledgements in the initial version of the paper

% If a paper is accepted, the final camera-ready version can (and usually should)
% include acknowledgements.  Such acknowledgements should be placed at the end of
% the section, in an unnumbered section that does not count towards the paper
% page limit. Typically, this will include thanks to reviewers who gave useful
% comments, to colleagues who contributed to the ideas, and to funding agencies
% and corporate sponsors that provided financial support.

\section*{Impact Statement}

This paper presents a new sparse attention framework aimed at improving the efficiency and accuracy of long-context inference in large language models. By reducing computational and memory overhead while preserving attention quality, this work contributes to more scalable and accessible deployment of LLMs in practical applications such as document understanding, retrieval, and reasoning over long inputs.

The techniques introduced in this paper are purely algorithmic and system-level improvements to existing attention mechanisms and do not introduce new capabilities that fundamentally alter the intended use of language models. As such, we do not foresee direct negative societal or ethical consequences arising uniquely from this work beyond those already associated with large language models more broadly.

% Authors are \textbf{required} to include a statement of the potential broader
% impact of their work, including its ethical aspects and future societal
% consequences. This statement should be in an unnumbered section at the end of
% the paper (co-located with Acknowledgements -- the two may appear in either
% order, but both must be before References), and does not count toward the paper
% page limit. In many cases, where the ethical impacts and expected societal
% implications are those that are well established when advancing the field of
% Machine Learning, substantial discussion is not required, and a simple
% statement such as the following will suffice:

% ``This paper presents work whose goal is to advance the field of Machine
% Learning. There are many potential societal consequences of our work, none
% which we feel must be specifically highlighted here.''

% The above statement can be used verbatim in such cases, but we encourage
% authors to think about whether there is content which does warrant further
% discussion, as this statement will be apparent if the paper is later flagged
% for ethics review.


\bibliography{example_paper}
\bibliographystyle{icml2026}

%%%%%%%%%%%%%%%%%%%%%%%%%%%%%%%%%%%%%%%%%%%%%%%%%%%%%%%%%%%%%%%%%%%%%%%%%%%%%%%
%%%%%%%%%%%%%%%%%%%%%%%%%%%%%%%%%%%%%%%%%%%%%%%%%%%%%%%%%%%%%%%%%%%%%%%%%%%%%%%
% APPENDIX
%%%%%%%%%%%%%%%%%%%%%%%%%%%%%%%%%%%%%%%%%%%%%%%%%%%%%%%%%%%%%%%%%%%%%%%%%%%%%%%
%%%%%%%%%%%%%%%%%%%%%%%%%%%%%%%%%%%%%%%%%%%%%%%%%%%%%%%%%%%%%%%%%%%%%%%%%%%%%%%
\newpage
\appendix
\onecolumn

\section{Full Benchmark Results}

\subsection{Ruler}

\begin{table}[h]
\centering
\small
\renewcommand{\arraystretch}{1.1}
\setlength{\tabcolsep}{3pt}
\caption{Ruler results across tasks for LLaMA-3.1-8B and Qwen3-8B, 16K context length. }
\begin{tabular}{lccccccccccccccc}
\hline
Method 
& s1\_niah 
& s2\_niah 
& s3\_niah 
& mk1\_niah 
& mk2\_niah
& mk3\_niah
& mv\_niah
& mq\_niah
& vt
& fwe
& cwe
& qa\_1
& qa\_2
& \textbf{Avg.} \\
\hline

LLaMA-3.1-8B             &100.00&100.00&100.00&100.00&100.00&100.00&96.50&100.00&99.60&89.60&98.00&82.00&64.00&94.59\\
Quest                    &99.00&100.00&87.00&100.00&96.00&70.00&95.25&97.75&98.60&86.50&88.00&84.00&58.00&89.24\\
RetroInfer               &100.00&99.50&99.50&100.00&98.50&96.50&95.62&99.88&99.90&82.40&87.00&76.00&55.50&91.56\\
Quest + Twi              &99.00&100.00&80.50&96.00&95.50&69.50&94.62&96.50&98.80&81.70&86.33&74.50&54.50&86.73\\
\textbf{Ours}            &100.00&100.00&100.00&100.00&99.5&98.00&97.00&99.75&99.50&92.90&88.67&78.00&54.00&92.87\\
\hline

Qwen3-8B                 &100.00&100.00&100.00&99.50&100.00&99.00&92.88&100.00&100.00&91.65&93.17&59.50&50.50&91.25\\
Quest                    &98.00&98.00&53.00&99.00&84.00&40.00&86.25&86.38&94.30&67.40&90.67&58.50&47.00&77.12\\
RetroInfer               &100.00&100.00&100.00&99.50&100.00&99.00&93.88&100.00&100.00&92.20&92.50&59.00&49.00&91.16\\
Quest + Twi              &99.00&98.50&52.50&98.50&83.50&36.50&86.88&86.25&92.70&66.20&90.00&61.00&45.50&76.69\\
\textbf{Ours}            &100.00&100.00&99.50&99.00&100.00&99.00&92.88&99.88&99.70&92.60&92.67&59.00&49.00&91.02\\
\hline

\end{tabular}
\label{tab:ruler_full_16}
\caption{Ruler results across tasks for LLaMA-3.1-8B and Qwen3-8B, 32K context length. }
\begin{tabular}{lccccccccccccccc}
\hline
Method 
& s1\_niah 
& s2\_niah 
& s3\_niah 
& mk1\_niah 
& mk2\_niah
& mk3\_niah
& mv\_niah
& mq\_niah
& vt
& fwe
& cwe
& qa\_1
& qa\_2
& \textbf{Avg.} \\
\hline

LLaMA-3.1-8B             &100.00 & 100.00 & 100.00 & 98.50 & 100.00 & 100.00 & 97.38 & 99.25 & 99.80 & 50.40 & 95.17 & 75.50 & 54.00 & 90.00 \\
Quest                    &97.50 & 100.00 & 86.50 & 99.00 & 97.50 & 89.00 & 92.38 & 96.50 & 99.00 & 44.50 & 90.00 & 73.00 & 52.50 & 85.95 \\
RetroInfer               &100.00 & 100.00 & 100.00 & 98.00 & 99.50 & 98.50 & 95.38 & 98.75 & 99.20 & 37.20 & 90.50 & 76.00 & 52.50 & 88.12 \\
Quest + Twi              &99.00 & 98.50 & 90.00 & 99.50 & 99.00 & 88.50 & 90.12 & 95.88 & 99.60 & 47.05 & 90.83 & 73.50 & 53.00 & 86.50 \\
\textbf{Ours}            &100.00 & 100.00 & 100.00 & 99.50 & 99.00 & 98.50 & 97.12 & 99.50 & 99.20 & 60.35 & 91.17 & 72.50 & 52.00 & 89.91 \\
\hline

Qwen3-8B                 &100.00 & 100.00 & 100.00 & 99.00 & 97.50 & 96.50 & 99.00 & 99.50 & 99.80 & 80.65 & 95.17 & 60.50 & 50.00 & 90.59 \\
Quest                    &100.00 & 100.00 & 72.50 & 99.50 & 89.00 & 43.50 & 93.12 & 86.62 & 92.60 & 60.85 & 90.17 & 62.00 & 47.50 & 79.80 \\
RetroInfer               &100.00 & 100.00 & 100.00 & 99.50 & 97.50 & 97.50 & 99.25 & 99.12 & 99.80 & 79.90 & 94.83 & 60.00 & 51.00 & 90.65 \\
Quest + Twi              &100.00 & 99.00 & 72.00 & 99.00 & 90.50 & 42.50 & 92.88 & 84.00 & 91.00 & 61.35 & 89.33 & 62.50 & 49.50 & 79.50 \\
\textbf{Ours}            &100.00 & 99.50 & 99.50 & 99.00 & 97.50 & 95.50 & 99.00 & 99.50 & 99.40 & 81.75 & 95.00 & 59.00 & 50.00 & 90.36 \\
\hline

\end{tabular}
\label{tab:ruler_full_32}
\caption{Ruler results across tasks for LLaMA-3.1-8B and Qwen3-8B, 64K context length. }
\begin{tabular}{lccccccccccccccc}
\hline
Method 
& s1\_niah 
& s2\_niah 
& s3\_niah 
& mk1\_niah 
& mk2\_niah
& mk3\_niah
& mv\_niah
& mq\_niah
& vt
& fwe
& cwe
& qa\_1
& qa\_2
& \textbf{Avg.} \\
\hline

LLaMA-3.1-8B             &100.00 & 100.00 & 100.00 & 100.00 & 99.00 & 96.50 & 98.38 & 99.75 & 98.70 & 9.50 & 86.33 & 72.50 & 49.00 & 85.36 \\
Quest                    &100.00 & 99.50 & 93.50 & 96.50 & 98.00 & 90.50 & 98.39 & 98.62 & 95.75 & 9.95 & 85.17 & 73.00 & 48.00 & 83.61 \\
RetroInfer               &100.00 & 100.00 & 100.00 & 100.00 & 96.50 & 90.50 & 97.50 & 99.75 & 98.10 & 5.05 & 79.67 & 72.50 & 49.50 & 83.77 \\
Quest + Twi              &100.00 & 99.50 & 93.00 & 96.00 & 95.50 & 71.50 & 99.25 & 96.25 & 96.70 & 11.10 & 85.00 & 72.50 & 48.00 & 81.87 \\
\textbf{Ours}            &100.00 & 100.00 & 100.00 & 100.00 & 98.00 & 92.50 & 97.62 & 99.75 & 96.90 & 13.75 & 79.67 & 72.50 & 48.50 & 84.55 \\
\hline

Qwen3-8B                 &100.00 & 100.00 & 100.00 & 90.50 & 76.50 & 31.00 & 95.50 & 97.25 & 99.70 & 68.20 & 78.50 & 66.50 & 41.50 & 80.40 \\
Quest                    &99.50 & 69.50 & 35.50 & 69.50 & 58.50 & 3.00 & 78.75 & 71.50 & 85.30 & 35.95 & 78.00 & 59.50 & 37.00 & 60.12 \\
RetroInfer               &100.00 & 99.50 & 100.00 & 90.00 & 75.00 & 29.50 & 94.88 & 97.12 & 99.70 & 68.75 & 77.67 & 67.50 & 41.50 & 80.09 \\
Quest + Twi              &99.50 & 62.00 & 31.50 & 80.00 & 63.64 & 0.00 & 73.38 & 65.62 & 80.43 & 31.43 & 75.83 & 57.72 & 40.68 & 58.59 \\
\textbf{Ours}            &100.00 & 98.50 & 96.50 & 89.00 & 72.00 & 28.00 & 94.50 & 95.25 & 97.00 & 70.50 & 78.67 & 69.00 & 41.50 & 79.26 \\
\hline

\end{tabular}
\label{tab:ruler_full_64}
\end{table}

\subsection{LongBench}

\begin{table}[h]
\centering
\small
\setlength{\tabcolsep}{3pt}
\renewcommand{\arraystretch}{1.1}
\caption{LongBench results across tasks for LLaMA-3.1-8B and Qwen3-8B.}
\begin{tabular}{lcccccccccccccc}
\hline
Method 
& 2Wiki
& Gov
& Hotpot 
& Lcc
& News
& Field 
& Musi
& Narr
& Retriv
& Qasper
& Qmsum
& Repo
& Triv
& \textbf{Avg.} \\
\hline

LLaMA-3.1-8B             &16.16 & 34.38 & 16.82 & 63.16 & 26.74 & 27.64 & 11.72 & 31.96 & 97.73 & 13.18 & 23.23 & 57.03 & 91.48 & 39.33 \\
Quest                    &16.27 & 34.51 & 16.69 & 63.22 & 26.86 & 27.27 & 11.94 & 33.77 & 96.58 & 12.94 & 23.75 & 50 & 90.03 & 38.76 \\
RetroInfer               &16.1 & 34.37 & 17.49 & 61.72 & 25.61 & 27.4 & 11.75 & 33.08 & 96.74 & 13.02 & 22.48 & 56.08 & 91.49 & 39.03 \\
Quest + Twilight         &16.38 & 34.05 & 16.72 & 63.55 & 26.82 & 26.92 & 11.54 & 31.43 & 95.78 & 12.85 & 23.25 & 56.23 & 91.02 & 38.96 \\
\textbf{Ours}            &16.18 & 34.93 & 17.1 & 62.85 & 26.78 & 27.36 & 11.47 & 32.07 & 96.22 & 12.67 & 22.66 & 56.17 & 91.32 & 39.06 \\
\hline

Qwen3-8B                 &12.68 & 29.87 & 12.86 & 67.52 & 23.04 & 26.38 & 7.5 & 3.74 & 95.46 & 11.44 & 21.12 & 63.79 & 90.21 & 35.82 \\
Quest                    &13.24 & 30.32 & 12.44 & 67.42 & 23.05 & 25.8 & 7.76 & 3.46 & 94.5 & 11.39 & 22.18 & 61.73 & 88.53 & 35.52 \\
RetroInfer               &12.08 & 29.91 & 12.73 & 67.5 & 22.69 & 26.04 & 7.89 & 3.62 & 94.25 & 11.43 & 20.53 & 64.35 & 90.21 & 35.63 \\
Quest + Twilight         &13.05 & 30.91 & 11.82 & 67.53 & 23.56 & 24.95 & 7.54 & 3.33 & 92.5 & 11.53 & 21.77 & 61.35 & 88.56 & 35.26 \\
\textbf{Ours}            & 12.61 & 30.02 & 12.48 & 67.71 & 22.16 & 26.27 & 7.54 & 3.74 & 95.21 & 11.72 & 20.77 & 64.48 & 90.46 & 35.78 \\
\hline

\end{tabular}
\label{tab:longbench_full}
\end{table}


% You can have as much text here as you want. The main body must be at most $8$
% pages long. For the final version, one more page can be added. If you want, you
% can use an appendix like this one.

% The $\mathtt{\backslash onecolumn}$ command above can be kept in place if you
% prefer a one-column appendix, or can be removed if you prefer a two-column
% appendix.  Apart from this possible change, the style (font size, spacing,
% margins, page numbering, etc.) should be kept the same as the main body.
%%%%%%%%%%%%%%%%%%%%%%%%%%%%%%%%%%%%%%%%%%%%%%%%%%%%%%%%%%%%%%%%%%%%%%%%%%%%%%%
%%%%%%%%%%%%%%%%%%%%%%%%%%%%%%%%%%%%%%%%%%%%%%%%%%%%%%%%%%%%%%%%%%%%%%%%%%%%%%%

\end{document}